\documentclass[runningheads]{llncs}
\usepackage{graphicx}

\usepackage{tikz}
\usepackage{comment}
\usepackage{amsmath,amssymb} 
\usepackage{color}

\usepackage{axessibility}  

\usepackage{multirow}
\usepackage{amssymb}
\usepackage{amsmath}
\usepackage{booktabs}
\usepackage{makecell}
\usepackage{subfigure}
\usepackage{marvosym}
\usepackage{caption}

\begin{document}
\pagestyle{headings}
\mainmatter
\def\ECCVSubNumber{1272}  

\title{FBNet: Feedback Network for Point Cloud Completion} 


\titlerunning{FBNet: Feedback Network for Point Cloud Completion}
%

\author{
Xuejun Yan\inst{1,3}$^{*}$  \and
Hongyu Yan\inst{2,1}\thanks{Equal contribution} \and
Jingjing Wang\inst{1} \and
Hang Du \inst{1} \and
Zhihong Wu \inst{2} \and
Di Xie \inst{1} \and
Shiliang Pu \inst{1}$^{(\textrm{\Letter})}$  \and
Li Lu \inst{2}$^{(\textrm{\Letter})}$
}
%
\authorrunning{Y. Xuejun et al.}


\institute{$^1$Hikvision Research Institute $^2$Sichuan University $^3$Zhejiang University\\
\email{\{yanxuejun,wangjingjing9,duhang,xiedi,pushiliang.hri\}@hikvision.com}\\
\email{hongyuyan@stu.scu.edu.cn, luli@scu.edu.cn}
}

\maketitle

\begin{abstract}
The rapid development of point cloud learning has driven point cloud completion into a new era. However, the information flows of most existing completion methods are solely feedforward, and high-level information is rarely reused to improve low-level feature learning. To this end, we propose a novel Feedback Network (\textbf{FBNet}) for point cloud completion, in which present features are efficiently refined by rerouting subsequent fine-grained ones.
Firstly, partial inputs are fed to a Hierarchical Graph-based Network (HGNet) to generate coarse shapes. Then, we cascade several Feedback-Aware Completion (FBAC) Blocks and unfold them across time recurrently. Feedback connections between two adjacent time steps exploit fine-grained features to improve present shape generations. The main challenge of building feedback connections is the dimension mismatching between present and subsequent features. To address this, the elaborately designed point Cross Transformer exploits efficient information from feedback features via cross attention strategy and then refines present features with the enhanced feedback features. Quantitative and qualitative experiments on several datasets demonstrate the superiority of proposed FBNet compared to state-of-the-art methods on point completion task. The source code and model are available at \url{https://github.com/hikvision-research/3DVision/}.

\keywords{Point cloud completion, Feedback Network, Cross transformer}
\end{abstract}

\section{Introduction}

With the rapid development of 3D sensors, point cloud has been widely used in 3D computer vision applications such as autonomous driving, augmented reality, and robotics. However, due to the limitations of the resolution, occlusion, and view angles, point clouds acquired from 3D sensors are usually sparse, partial, and noisy. Therefore, recovering the complete shape from its partial observation is desirable and vital for various downstream tasks (e.g. shape classification~\cite{qi2017pointnet}, object detection~\cite{Li2019GS3DAE,Qi2019DeepHV}, and semantic/instance segmentation~\cite{wen2020point}).

\begin{figure}[t]
\centering
\scalebox{0.8}{
\includegraphics[width=\textwidth]{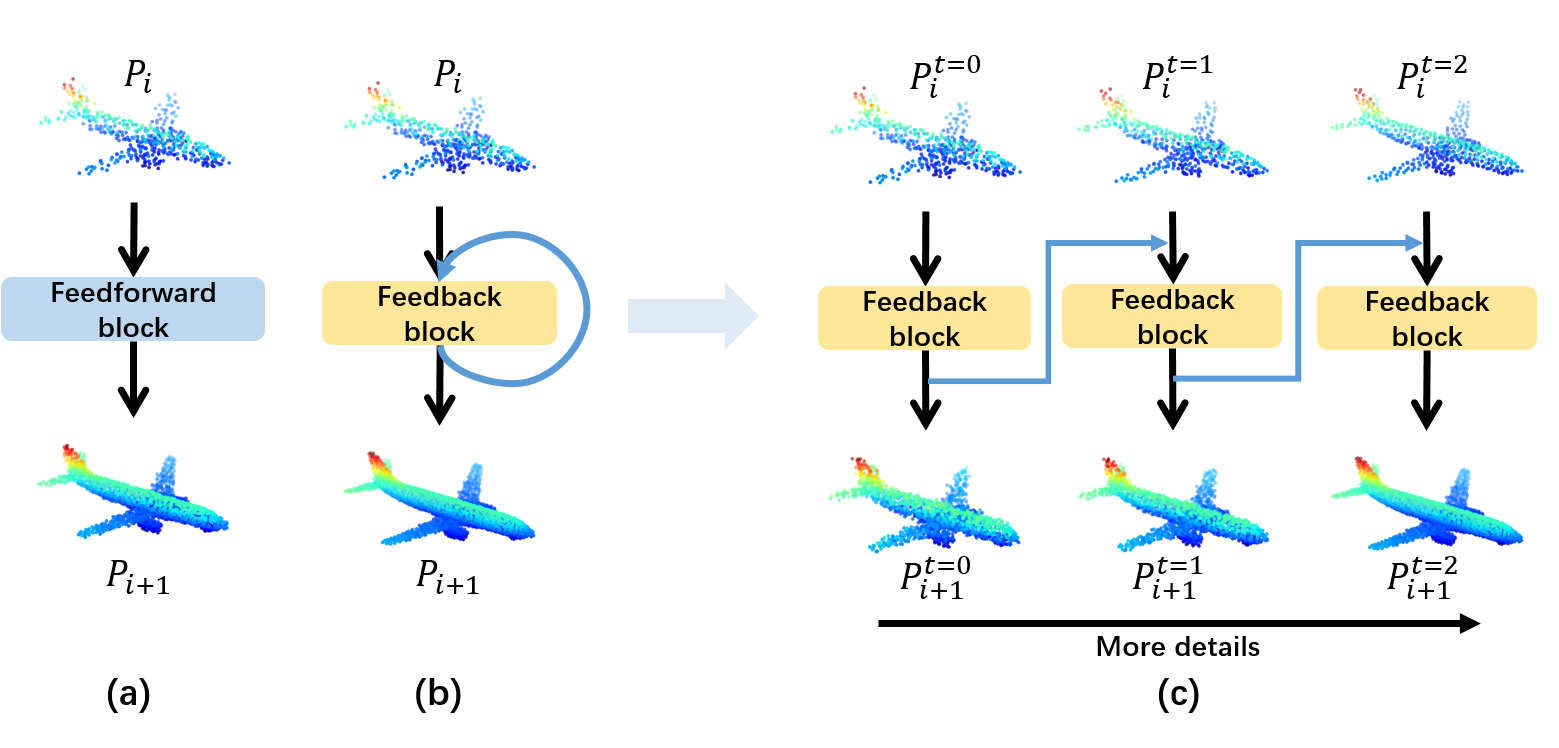}}
\caption{The illustrations of the feedforward and feedback blocks for point completion. (a) Feedforward block in which information only flows from low-level block to high-level block. (b) Feedback mechanism in our proposed FBNet which refines present information with high-level information via feedback connections (represented by the blue arrow). (c) Unfolding feedback block across time steps.}
\label{fig:feedback-intro}
\end{figure}

The pioneering work PCN \cite{yuan2018pcn} applied an encoder-decoder net to predict coarse shapes from partial inputs firstly, and then refined coarse results to fine-grained completions through a folding-based decoder. Enlightened by the success of PCN's two-stage generation fashion (coarse-to-fine), most recent point completion works~\cite{yuan2018pcn,tchapmi2019topnet,wang2020cascaded,liu2020morphing,wang2020softpoolnet,pan2021variational} focus on enhancing the the performance of the network of coarse or fine stages to get more fruitful final results. More recent works~\cite{wang2020cascaded,xiang2021snowflakenet,huang2021rfnet} extended two-stage (coarse-to-fine)  strategy to multistage point generation and achieved impressive completion performance. However, these completion networks are wholly feedforward: the information solely flows from lower stages to higher ones. It is reasonable to infer that the output of a regular completion network has more details and higher quality compared to its input, as shown in Fig.~\ref{fig:feedback-intro} (a). Similarly, for a two-stage or multistage network, higher stages also generate better results compared to lower ones. \textsl{Is it possible to utilize higher quality information to make the lower stage features more representative and informative?}

 Motivated by the successful applications of the feedback mechanism on 2D vision~\cite{Li2019FeedbackNF,Sam2018TopDownFF,Feng2019AttentiveFN,Zeng2020HighResolutionII,Chen2021RobustRL,Li2019GatedMF}, we propose the feedback network (FBNet) for point cloud completion. To the best of our knowledge, FBNet is the first feedback-based network for point completion task. The feedback mechanism aims to transmit high-level features to previous layers and refine low-level information.
As shown in Fig.~\ref{fig:feedback-intro} (b), the feedback block reuses the output information to enhance its own feature learning via feedback connections. When unfolding it across time, the feedback block takes the state of the previous time step to enrich present feature representations, shown in Fig.~\ref{fig:feedback-intro} (c). As a result, the feedback block has the ability to refine its output recurrently across time steps.

The proposed FBNet consists of a Hierarchical Graph-based Network (HGNet) and the feedback refinement module which stacks three Feedback-Aware Completion (FBAC) blocks, as shown in Fig.~\ref{fig:overview}. The HGNet is an encoder-decoder structure that encodes partial inputs to global features and then decodes them to coarse complete point clouds. The existing point cloud pooling methods~\cite{qi2017pointnet++,zhou2021adaptive} downsample point clouds via FPS algorithm and focus on pooled point feature learning. Due to unlearnability, the FPS algorithm may fail to correctly capture local structures when the input is partial and sparse. Hence, the Adaptive Graph Pooling (AdaptGP), the learnable downsampling method, is applied in the HGNet to get more fruitful and accurate global features.
After getting coarse complete shapes from HGNet, the feedback refinement module with stacked FBAC blocks refine the coarse complete point clouds to fine-grained and dense ones.
For each FBAC block, the high-level information is rerouted to the low layer through the feedback connection and helps to make low-layer features more representative and informative. The main challenge of fusing high-layer information with low-layer ones is the dimension mismatching between two layers.
To address this problem, the proposed Cross Transformer builds the relationships between two mismatched point features, and adaptively selects useful information from rerouted features to enhance low-layer features via cross attention strategy. The stacked FBAC blocks gradually refine their outputs across time steps and finally get impressive complete shapes.
The proposed FBNet achieves state-of-the-art performance on several benchmarks with various resolutions. Our key contributions are as follows:
\begin{itemize}
\item We propose a novel feedback network (FBNet) for point completion, which recurrently refines completion shapes across time steps. To the best of our knowledge, the proposed FBNet is the first feedback-based network for point completion task.
\item We introduce the feedback-aware completion (FBAC) block to refine coarse complete point clouds to fine-grained ones. Compared with previous feedforward completion works, feedback connections in FBAC blocks reroute high-level point information back to the low layer and make low-level features more representative.
\item We design the Cross Transformer to overcome the feature mismatching problem and adaptively select valuable information from feedback features for enhancing low-level features.
\item Experiments on several datasets show that the proposed FBNet achieves superior performance compared to state-of-the-art methods.
\end{itemize}

\section{Related Work}

\subsection{Point Cloud Processing}

The pioneering work PointNet~\cite{qi2017pointnet} used the shared Multi-layer Perceptions (MLPs) and symmetrical max-pooling operation to extract global features on the unorderedness point cloud, which did not take the relationships of local points into count.
To solve this problem, PointNet++ \cite{qi2017pointnet++} introduced a hierarchical architecture with local PointNet layers to capture regional information. Following point-based works~\cite{li2018so,gadelha2018multiresolution,yan2020pointasnl,guo2021pct,ran2021learning,zhao2021point,xiang2021walk,ran2022surface} focused on how to learn local features more effectively. Recently, PCT~\cite{guo2021pct} and Point Transformer~\cite{zhao2021point} introduced the self-attention strategy of Transformer~\cite{vaswani2017attention} for point feature learning and achieved impressive results. For achieving convolution-like operation in the point cloud, many convolution-based works~\cite{dai2017deformable,li2018pointcnn,xu2018spidercnn,liu2019relation,wu2019pointconv,thomas2019kpconv,xu2021paconv} built the relationship between local centers and their neighborhoods to learn dynamic weights for convolution. Besides, graph-based methods~\cite{kipf2016semi,wang2019dynamic,xu2020grid,zhou2021adaptive} achieved notable performance for the local aggregation of geometric features, where DGCNN~\cite{wang2019dynamic} proposed the EdgeConv module to dynamically update point features by learning edge features between two nodes. In this paper, we aim to design a hierarchical graph-based encoder to learn multi-scale geometric features of the partial point cloud.


\subsection{Point Cloud Completion}

The target of point cloud completion is to recover a complete 3D shape based on its partial observation. Recent advances based on 3D point cloud processing techniques have boosted the research of point cloud completion. PCN~\cite{yuan2018pcn} first proposed an explicit coarse-to-fine completion framework, which generates a coarse point cloud based on learned global features from the partial inputs and then refines coarse results to fine-grained completions through a folding-based decoder. Enlightened by the success of PCN, the following methods~\cite{tchapmi2019topnet,xie2020grnet,liu2020morphing,wen2020point,wang2020softpoolnet,huang2020pf,zhang2020detail,wang2020cascaded,wen2021cycle4completion,wen2021pmp,yu2021pointr,huang2021rfnet} especially focused on the local feature exploitation, and the decoding operation were applied to refine their 3D completion results. Most recently, VRCNet~\cite{pan2021variational} introduced a novel variational relation point completion dual-path network, which used the probabilistic modeling network and relation enhancement network to capture local point features and structural relations between points. SnowflakeNet~\cite{xiang2021snowflakenet} stacked several SPD modules to generate a complete shape like the snowflake growth of points in 3D space. Although these methods achieved impressive completion performance, all of them are feedforward networks that the information flows only from low layer to high layer, and ignore that the information from fine-grained shapes can reroute back to correct previous states.

\subsection{Feedback mechanism}
The feedback mechanism has been widely employed in various 2D image vision tasks~\cite{Li2019FeedbackNF,Sam2018TopDownFF,Feng2019AttentiveFN,Zeng2020HighResolutionII,Chen2021RobustRL,Li2019GatedMF}. With feedback connections, high-level features are rerouted to the low layer to refine low-level feature representations. Building feedback connections in 2D vision tasks are convenient because image-based feature maps are regular and ordered. The resolutions of high-layer feature maps can strictly align with lower ones easily. Several works~\cite{Li2019FeedbackNF,Li2019GatedMF,Zeng2020HighResolutionII} directly concatenate feedback features with lower ones and then employ convolution layers to fuse features. Similarly, in our work, the high-resolution point features are transmitted back to enrich low-resolution point features. Due to the unorderedness of the point cloud, it is difficult to align two point cloud features with different resolutions.



\section{Methodology}

\begin{figure}[t]
\centering
\scalebox{0.9}{
\includegraphics[width=\textwidth]{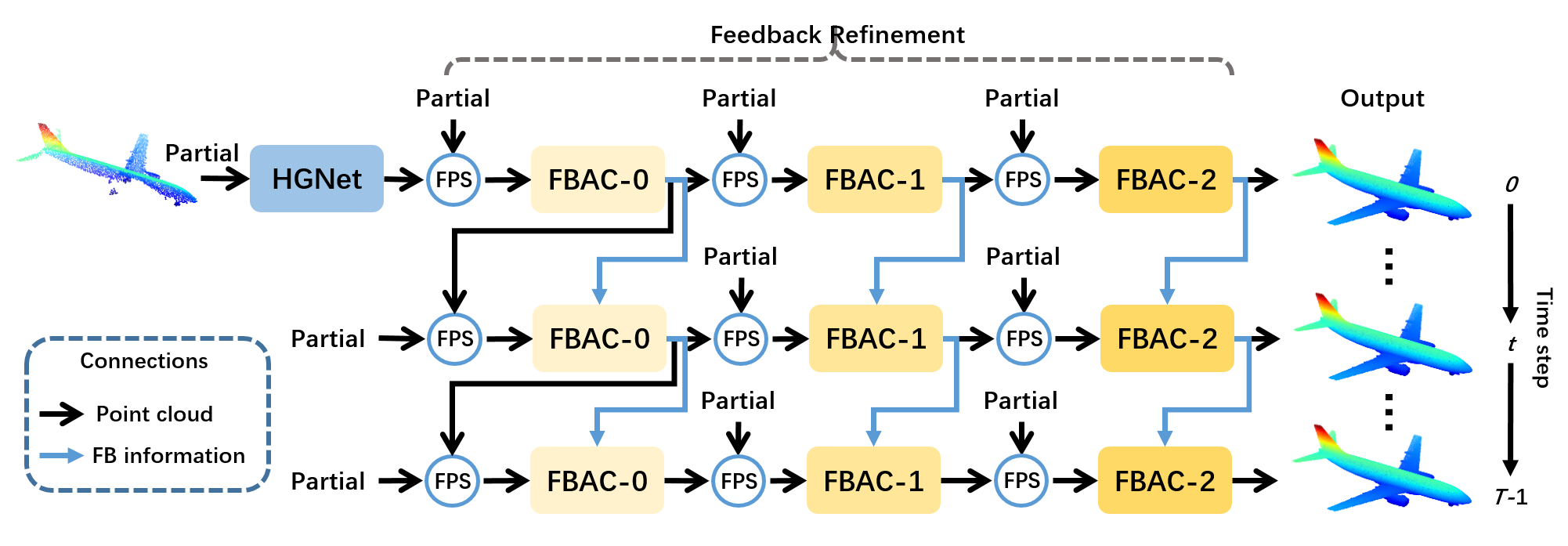}}
\caption{The overall architecture of our FBNet consists of the Hierarchical Graph-based Network (HGNet) and the feedback refinement module that stacks three Feedback-Aware Completion (FBAC) Blocks. The HGNet aims to generate coarse completions from partial inputs. The cascaded FBAC blocks in the feedback refinement module aim to reconstruct complete and dense point clouds from partial inputs and coarse outputs. Note that, the FBAC block's  weight parameters are shared across time steps.}
\label{fig:overview}
\end{figure}

The overall architecture of FBNet is shown in Fig.~\ref{fig:overview}, which consists of two modules: one Hierarchical Graph-based Network (HGNet) and the feedback refinement module with three Feedback-Aware Completion (FBAC) blocks. The HGNet generates sparse but complete shapes from partial inputs. The stacked FBAC blocks reconstruct complete and fine-grained point point clouds based on both partial inputs and HGNet's outputs. The feedback connections on these FBAC blocks reroute high resolution point information (points and features) to enrich low resolution point features. With the help of feedback mechanism, FBAC blocks can gradually refine their outputs across time steps and finally get impressive complete shapes.

\subsection{Hierarchical Graph-based Network}

\begin{figure}[t]
\centering
\scalebox{0.5}{
\includegraphics[width=\textwidth]{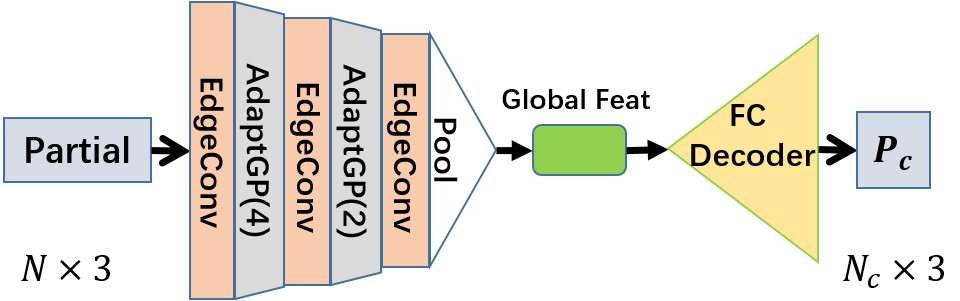}}
\caption{The architecture of Hierarchical Graph-based Network.}
\label{fig:HGNet}
\end{figure}

Hierarchical Graph-based Network (HGNet) is an encoder-decoder structure, which encodes partial inputs to global features and then decodes them to coarse complete point clouds, as shown in Fig.~\ref{fig:HGNet}. To get more representative global features, we introduce the hierarchical encoder which stacks EdgeConv~\cite{wang2019dynamic} and proposed Adaptive Graph Pooling alternately.

{\bf Adaptive Graph Pooling.} The existing point
cloud pooling methods~\cite{qi2017pointnet++,zhou2021adaptive} usually downsampled point clouds via FPS algorithm and aggregated sampled point feature by pooling their neighborhood features. However, the FPS algorithm is unstable, which may fail to pool geometric information efficiently when the input is incomplete and sparse.

To overcome above drawback, we propose a novel Adaptive Graph Pooling (AdaptGP) to pool points and features for partial inputs adaptively. Specifically, AdaptGP generates the pooled points and features by weighting the neighbors of sampled points in geometric and feature space respectively. The weight matrix $w$ is learned from the relation of key point and its neighbors, which can be defined as:
\begin{equation}
w = Softmax(\mathcal{M}((f_{i}-f_{j})+\mathcal{K}(p_{i} - p_{j}))
\end{equation}
where $\mathcal{M}$, $\mathcal{K}$ are linear mapping functions (i.e., MLPs), which can learn both the point and feature relations between point $p_{i}$ and point $p_j$.

The final pooled point and feature of in our AdaptGP can be represented as:
\begin{eqnarray}
p_{i}^{\prime}=\sum _{j:(i, j) \in \mathcal{E}} w_{j}p_{j} \\
f_{i}^{\prime}=\sum _{j:(i, j) \in \mathcal{E}} w_{f,j}f_{j}
\end{eqnarray}

where $\mathcal{E} \subseteq \mathcal{V} \times \mathcal{V}$ is the edge set of the directed point graph $G(\mathcal{V}, \mathcal{E})$ built by k-nearest neighbors(kNN) grouping operation.
In our encoder, we use the two AdaptGP with the pooling rate (4, 2) and three EdgeConv with dimensions (64, 128, 512) to extract multi-scale point features. Finally, we employ max and avg pooling operations to generate the global feature.

{\bf Coarse Point Generator.} We decode the global feature to predict a coarse output $P_{c}$ via three fully-connected (FC) layers, as shown in Fig.~\ref{fig:HGNet}.

With the help of the representative global feature generated by our hierarchical graph-based encoder, the FC layers can recover the coarse shape with more geometric details. Besides, we use FPS to gain new coarse output $P_{c}^{\prime}$ from aggregation of $P_{c}$ and partial input, which provides a better initial input for the following feedback refinement stage.

\begin{figure}[t]
\centering
\scalebox{0.9}{
\includegraphics[width=\textwidth]{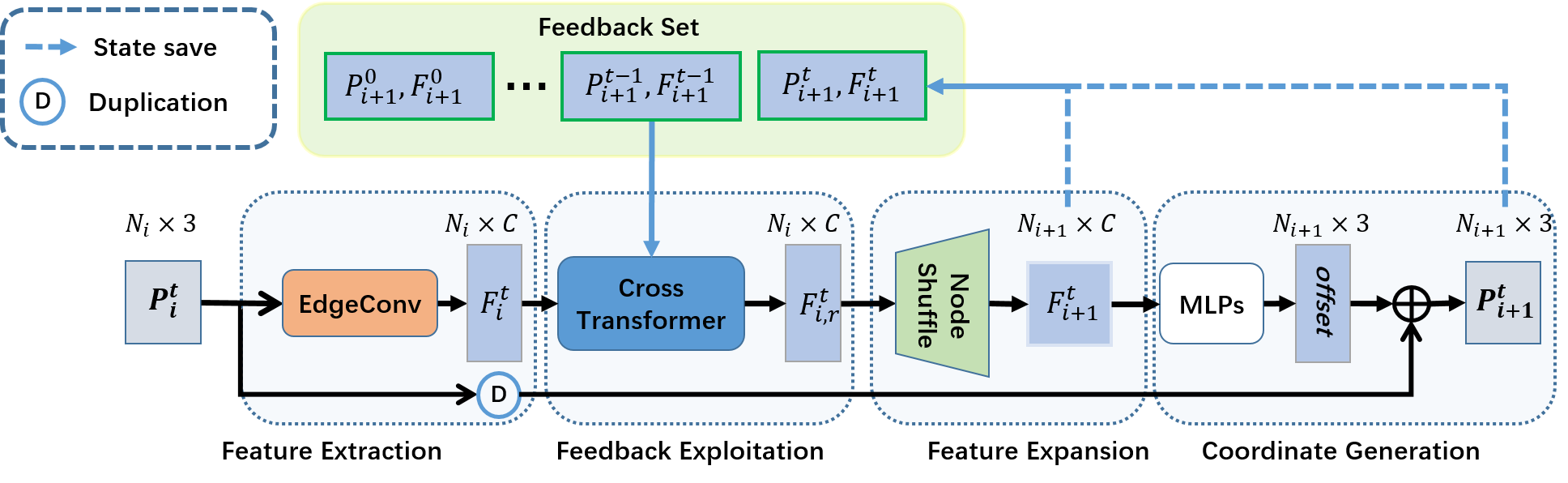}}
\caption{The detailed architecture of proposed Feedback-Aware Completion Block which consists of four parts: feature extraction, feedback exploitation, feature expansion and coordinate generation.}
\label{fig:FBAC}
\end{figure}

\subsection{Feedback Refinement Mechanism}

The feedback refinement module aims to refine the HGNet's coarse result to the complete and fine-grained point cloud. The feedback refinement module consists of three stacked Feedback-Aware Completion (FBAC) blocks and builds multiple feedback connections at various scales to learn more effective features.
As shown in Fig.~\ref{fig:overview} (left-to-right), the information firstly flows from coarse HGNet to stacked FBAC blocks in a feedforward fashion.
Each FBAC block takes the output of previous block as input, then refines and upsamples the shape to the denser one with more details.

In particular, for getting better input to $i$-th FBAC block, we aggregate partial input with feedforward output of $i-1$-th block to a new point cloud, and then downsample it to the same size of previous output via FPS algorithm. Through this initialization operation, the $i-1$-th block's outputs are refined
with the original geometric information from partial shapes to get the new inputs, which make present block easier to generate fruitful results.
For $0$-th block, we aggregate partial input with HGNet's output to initialize its input at $t=0$ step, and aggregate partial input with its own output at $t-1$ step when $t>0$.

In our refinement module, the information also flows from high layer to low layer of same FBAC block in the feedback fashion. We unfold FBAC blocks across time steps, shown in Fig.~\ref{fig:overview} (top-to-down). The weight parameters of FBAC blocks are shared across time steps. For the $i$-th FBAC block at $t$ step, its high layer features at $t-1$ step are rerouted and used for present step feature learning via the feedback connection. It is reasonable to infer that high layer features at $t-1$ step contain fine-grained information that can refine the low layer features to be more representative at present time $t$ step. As a result, stacked FBAC blocks gradually refine their outputs across time steps and finally fruitful shapes are generated via the feedback refinement mechanism.


\subsection{Feedback-Aware Completion Block.}

The Feedback-Aware Completion (FBAC) block aims to refine and upsample the low resolution point cloud to high resolution and fine-grained ones via aggregating the points and features from the feedback connection. As shown in Fig.~\ref{fig:FBAC}, the detailed architecture of proposed FBAC block consists of four parts: feature extraction, feedback exploitation, feature expansion and coordinate generation. We first use EdgeConv~\cite{wang2019dynamic} to extract local geometric features $F_{i}^{t}$ from $P_{i}$. Then, the Cross Transformer fuses present features $F_{i}^{t}$ with feedback information $(P_{i+1}^{t-1},F_{i+1}^{t-1})$ generated at $t-1$ step. Subsequently, the refined features $F_{i,r}^{t}$ is expanded $r$ times via the NodeShuffle~\cite{Qian2021PUGCNPC} block, where $r=N_{i+1}/N_{i}$. Finally, we gain the point displacements via a series of MLPs.

The final result is calculated as:

\begin{equation}
P_{i+1}^{t} = \left\{\mathcal{D}(P_{i}^t,r) +\mathcal{M}(\mathcal{\eta}(\mathcal{T}(P_{i}^{t},\mathcal{G}(P_{i}^{t}), P_{i+1}^{t-1}, F_{i+1}^{t-1}), r))\right\}
\end{equation}
where $\mathcal{D}$ is the duplication operation. $\mathcal{M}$ is a series of linear functions. $\eta$ is node-shuffle operation and $r$ is up-sampling ratio. $\mathcal{T}$ is Cross Transformer function. $\mathcal{G}$ is the feature extractor based on EdgeConv.


\begin{figure}[t]
\centering
\scalebox{0.75}{
\includegraphics[width=\textwidth]{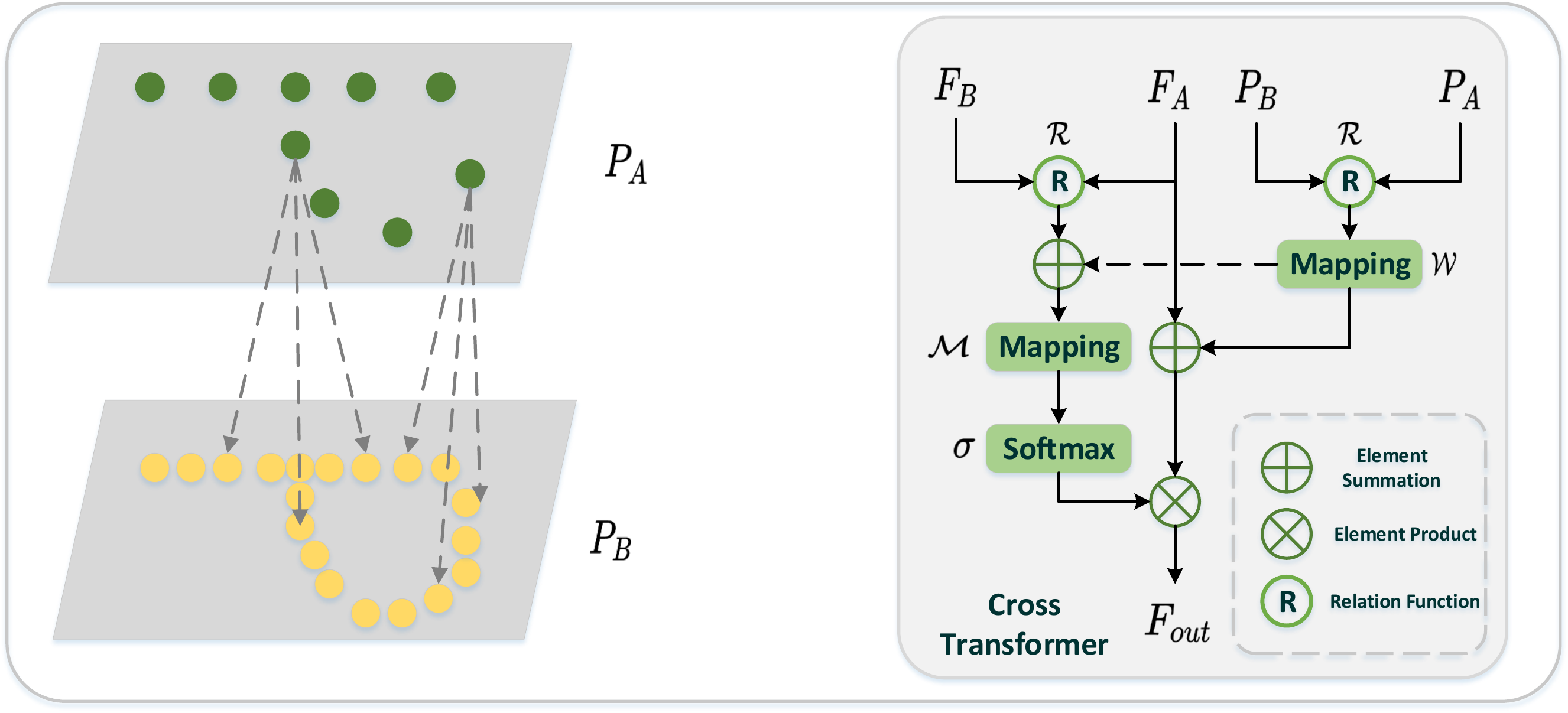}}
\caption{The detailed structure of our Point Cross Transformer. The left figure visualize the perception process (noted as dashed arrows) of Cross Transformer between two point clouds $P_{A}$,$P_{B}$. The detail structure of Cross Transformer is shown on the right. $P_{A}$, $F_{A}$ and $P_{B}$, $F_{B}$ denote two different point clouds and their features. The $\mathcal{M}$ and $\mathcal{W}$ are linear mapping functions. $\mathcal{R}$ is a relation function (i.e., subtraction)}
\label{fig:cross}
\end{figure}
{\bf Cross Transformer.} The cross layer feature fusion strategy is widely used for point completion methods~\cite{wen2020point,pan2021variational,xiang2021snowflakenet}. But in most of them, low-level information flows from a low layer to a high layer with same resolution in a feedforward fashion and the alignment between two features is required. Take the skip-transformer in SnowflakeNet~\cite{xiang2021snowflakenet} for example, the two skip-connected features are the displacement feature of the previous SPD and the points feature extracted from the input of the present SPD (also the output of the previous SPD). These two features have the same resolution and a strict one-to-one mapping is built between them. As a result, it is convenient to fuse them via the concatenation operation in the Skip Transformer~\cite{xiang2021snowflakenet}.


In our FBAC block, the feedback feature $F_{i+1}^{t-1}$ has higher resolution compared with the extracted feature $F_{i}^{t}$. Regular fusion strategies (e.g., Skip Transformer, Concatenation) can not be applied directly since they only process the fusion of two point clouds that have the same resolution and their features are aligned.


To address this, we propose a novel point Cross Transformer to fuse features from two point clouds with different resolutions. As shown in Fig.~\ref{fig:cross}, the $P_{A}$ and $P_{B}$ are two point clouds with different resolutions. By constructing cross-points relations between $P_{A}$ and $P_{B}$, our Cross Transformer can utilize $P_{B}$ information to guide the network to learn better feature representation of $P_{A}$. The calculation of Cross Transformer can be formulated as:
\begin{equation}
f_{a}^{\prime}=\sum _{b:(a, b) \in \mathcal{E(A,B)}} \sigma(\mathcal{M}\left(f_{a}-f_{b}+\delta\right)) \odot (f_{b}+\delta)
\end{equation}
where $\mathcal{E(A,B)}$ is the directed edge set from point cloud $P_{A}$ to $P_{B}$. the $\mathcal{M}$ is a mapping function i.e., $\{MLP \rightarrow ReLU \rightarrow MLP \}$. The $\delta = \mathcal{W}(p_{a}-p_{b})$ which encoding the position embedding. The $\sigma$ denotes $Softmax$ function. Through building the cross point cloud graph, the proposed Cross Transformer has the ability to query useful information from reference point cloud $P_{B}$ and then enrich its own features, shown in Fig.~\ref{fig:cross}. When $P_{A},P_{B}$ are the same point cloud, our Cross Transformer degrades to a point transformer~\cite{zhao2021point} with self-attention mechanism.

The Cross Transformer refines present point features with the reference point cloud information via the cross attention strategy. We apply the Cross Transformer to exploit feedback information and enrich present feature representations. In our FBAC block, we build $P_{A},P_{B}$ and their features as follows:
\begin{eqnarray}
P_{A}&=&P_{i}^{t}, \quad F_{A}=F_{i}^{t} \nonumber \\
P_{B}&=&[P_{i}^{t}, P_{i+1}^{t-1}], \quad F_{B}=[F_{i}^{t}, F_{i+1}^{t-1}]
\end{eqnarray}
where $[,]$is the merge operation.

As a result, present features $F_{i}^{t}$ adaptively query valuable information from the merged feature set. It should be noted that there is no feedback features at $0$ step , the $F_{i}^{t}$ queries its own features and Cross Transformer degenerate to a Point Transformer.

\subsection{Training Loss}

In our implementation, we use Chamfer Distance (CD) as the loss function, which can be defined as follows:

\begin{equation}
\mathcal{L}_{CD}\left({P_{1}},{P_{2}}\right)=\frac{1}{\left|{P_{1}}\right|} \sum_{x \in {P_{1}}} \min _{y \in {P_{2}}}\|x-y\|^{2} +\frac{1}{\left|{P_{2}}\right|} \sum_{y \in {P_{2}}} \min _{x \in {P_{1}}}\|y-x\|^{2}.
\end{equation}
where $x$ and $y$ denote points that belong to two point clouds ${P_{1}}$ and ${P_{2}}$, respectively.

The total training loss is formulated as:
\begin{equation}
\mathcal{L}=\mathcal{L}_{CD}(P_{c}, Y_{gt})+\sum_{t=0}^{T}\sum_{i=1}^{n}\mathcal{L}_{CD}(P_{i}^{t}, Y_{gt}),
\end{equation}
where $P_{c}$ denotes the coarse output of HGNet. The $P_{i}^{t}$ and $Y_{gt}$ denote the output of $i-1$-th FBAC block at $t$ step and ground truth, respectively. $T$ is the number of the time steps.

\section{Experiments}

{\bf Implementation and Evaluation Metrics.} The number of time steps in our FBNet is set to 3 for low resolution (2K) task and for other higher resolution tasks (e.g., 4K, 8K, 16K) is set to 2. Full settings are detailed in the supplementary material. The model's performance is evaluated by Chamfer Distance (CD) and F1-score.


\begin{table}[t]
\scriptsize
\begin{center}
\caption{Point cloud completion results on MVP dataset (16384 points) in terms of per-point L2 Chamfer distance  ($\times 10^{4}$). The proposed  FBNet achieves
the lowest reconstruction errors in 14 categories. The best results are highlighted in bold.}
\scalebox{0.9}{
\begin{tabular}{l|cccccccccccccccc|c}
\Xhline{2\arrayrulewidth}
Methods & \rotatebox{80}{airplane} & \rotatebox{80}{cabinet} &  \rotatebox{80}{car}  &  \rotatebox{80}{chair} & \rotatebox{80}{lamp} & \rotatebox{80}{sofa} & \rotatebox{80}{table} & \rotatebox{80}{watercraft} & \rotatebox{80}{bed} & \rotatebox{80}{bench} & \rotatebox{80}{bookshelf} & \rotatebox{80}{bus} & \rotatebox{80}{guitar} & \rotatebox{80}{motorbike} & \rotatebox{80}{pistol} & \rotatebox{80}{skateboard} & Avg.  \\
\hline
PCN \cite{yuan2018pcn} & 2.95 & 4.13 & 3.04 & 7.07 & 14.93 & 5.56 & 7.06 & 6.08 & 12.72 & 5.73 & 6.91 & 2.46 & 1.02 & 3.53 & 3.28 & 2.99 & 6.02 \\
TopNet \cite{tchapmi2019topnet} & 2.72 & 4.25 & 3.40 & 7.95 & 17.01 & 6.04 & 7.42 & 6.04 & 11.60 & 5.62 & 8.22 & 2.37 & 1.33 & 3.90 & 3.97 & 2.09 & 6.36 \\
MSN \cite{liu2020morphing} & 2.07 & 3.82 & 2.76 & 6.21 & 12.72 & 4.74 & 5.32 & 4.80 & 9.93 & 3.89 & 5.85 & 2.12 & 0.69 & 2.48 & 2.91 & 1.58 & 4.90 \\
CRN \cite{wang2020cascaded} & 1.59 & 3.64 & 2.60 & 5.24 & 9.02 & 4.42 & 5.45 & 4.26 & 9.56 & 3.67 & 5.34 & 2.23 & 0.79 & 2.23 & 2.86 & 2.13 & 4.30 \\
GRNet \cite{xie2020grnet} & 1.61 & 4.66 & 3.10 & 4.72 & 5.66 & 4.61 & 4.85 & 3.53 & 7.82 & 2.96 & 4.58 & 2.97 & 1.28 & 2.24 & 2.11 & 1.61 & 3.87 \\
NSFA \cite{zhang2020detail} & 1.51 & 4.24 & 2.75 & 4.68 & 6.04 & 4.29 & 4.84 & 3.02 & 7.93 & 3.87 & 5.99 & 2.21 & 0.78 & 1.73 & 2.04 & 2.14 & 3.77 \\
VRCNet \cite{pan2021variational} & 1.15 & 3.20 & \textbf{2.14} & 3.58 & 5.57 & 3.58 & 4.17 & 2.47 & 6.90 & 2.76 & 3.45 & 1.78 & 0.59 & \textbf{1.52} & 1.83 & 1.57 & 3.06 \\
SnowflakeNet \cite{xiang2021snowflakenet} & 0.96 & 3.19 & 2.27 & 3.30 & 4.10 & 3.11 & 3.43 & 2.29 & 5.93 & 2.29 & 3.34 & 1.81 & 0.50 & 1.72 & 1.54 & 2.13 & 2.73  \\
\hline
 \textbf{FBNet (Ours)} & \textbf{0.81} & \textbf{2.97} & 2.18 & \textbf{2.83} & \textbf{2.77} & \textbf{2.86} & \textbf{2.84} & \textbf{1.94} & \textbf{4.81} & \textbf{1.94} & \textbf{2.91} & \textbf{1.67} & \textbf{0.40} & 1.53 & \textbf{1.29} & \textbf{1.09} & \textbf{2.29} \\
\Xhline{2\arrayrulewidth}
\end{tabular}}
\label{tab:mvp}
\end{center}
\end{table}

\begin{table}[t]
\begin{center}
\caption{Quantitative results with various resolutions on the MVP dataset. For CD ($\times 10^{4}$), lower is better. For F1-score, higher is better.}
\scalebox{0.9}{
\begin{tabular}{l|cc|cc|cc|cc}
\Xhline{2\arrayrulewidth}
\multirow{2}{*}{Method} & \multicolumn{2}{c|} {2048} & \multicolumn{2}{c|} {4096} & \multicolumn{2}{c|} {8192} & \multicolumn{2}{c} {16384} \\ \cline{2-9}
& CD & F1 & CD & F1 & CD & F1 & CD & F1 \\
\hline
\hline
PCN~\cite{yuan2018pcn}  & 9.77 & 0.320 & 7.96 & 0.458 & 6.99 & 0.563 & 6.02 & 0.638 \\
TopNet~\cite{tchapmi2019topnet}  & 10.11 & 0.308 & 8.20 & 0.440 & 7.00 & 0.533 & 6.36 & 0.601 \\
MSN~\cite{liu2020morphing} & 7.90 & 0.432 & 6.17 & 0.585 & 5.42 & 0.659 & 4.90 & 0.710 \\
CRN~\cite{wang2020cascaded}  & 7.25 & 0.434 & 5.83 & 0.569 & 4.90 & 0.680 & 4.30 & 0.740 \\
VRCNet~\cite{pan2021variational} & 5.96 & 0.499 & 4.70 & 0.636 & 3.64 & 0.727 & 3.12 & 0.791\\
SnowflakeNet~\cite{xiang2021snowflakenet} & 5.71 & 0.503 & 4.40 & 0.661 & 3.48 & 0.743 & 2.73 & 0.796\\
\hline
FBNet \textbf{(Ours)} & \textbf{5.06} & \textbf{0.532} & \textbf{3.88} & \textbf{0.671} & \textbf{2.99} & \textbf{0.766} & \textbf{2.29} & \textbf{0.822}\\
\Xhline{2\arrayrulewidth}
\end{tabular}}
\label{tab:mvp_all}
\end{center}
\end{table}

\begin{figure}[t]
\centering
\scalebox{0.9}{
\includegraphics[width=\textwidth]{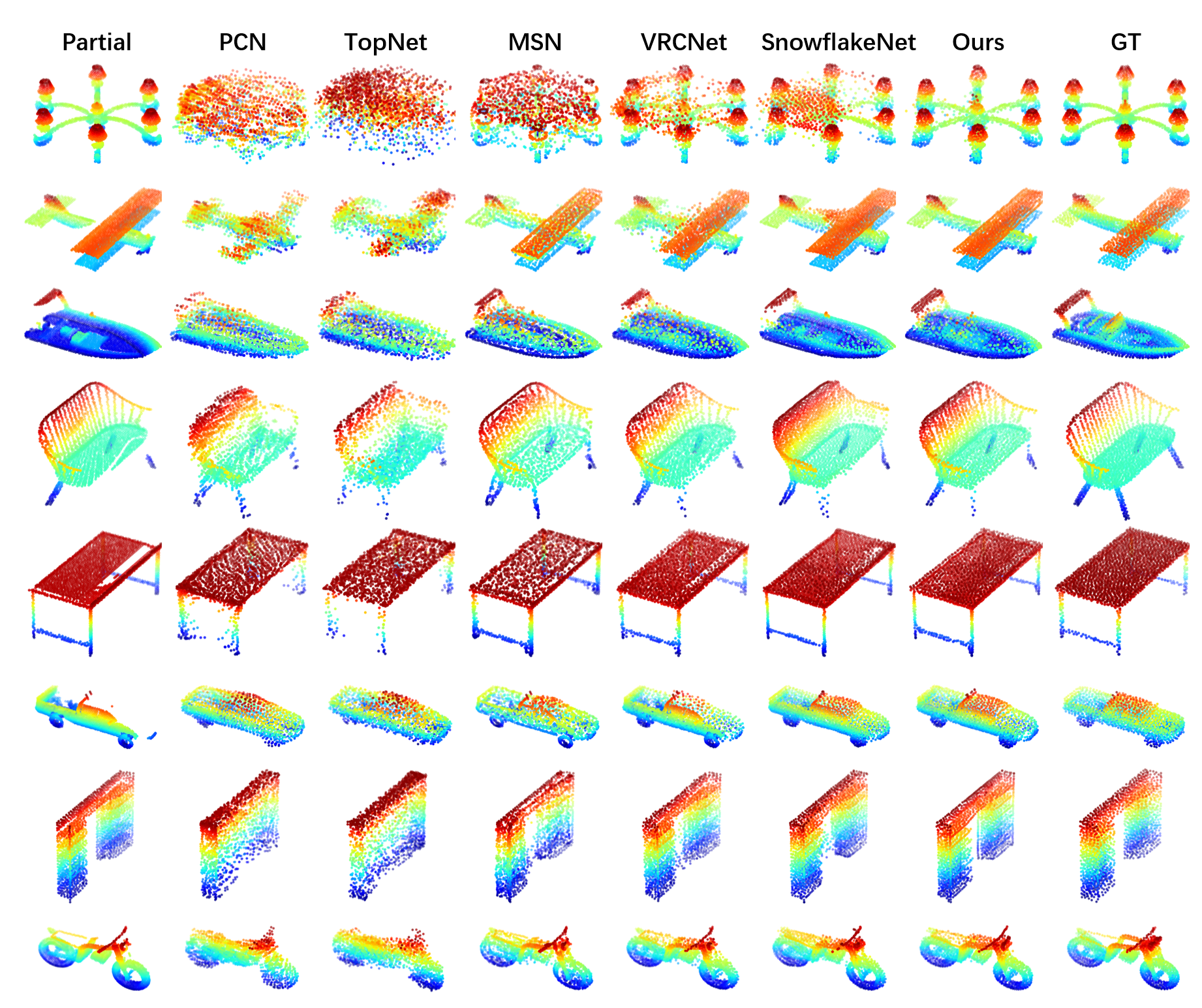}}
\caption{Qualitative comparison of different methods on MVP dataset (2048 points). Our FBNet generates better complete shapes with fine-grained details compared to other methods.}
\label{fig:mvp2k}
\end{figure}

\subsection{Evaluation on MVP Dataset.}
The MVP dataset is a multi-view partial point cloud dataset.
There are 26 incomplete shapes that
are captured from 26 uniformly distributed views for each CAD model.
Furthermore, MVP consists of 16 shape categories of partial and complete shapes for training and testing. MVP dataset provides various resolutions (including 2048, 4096, 8192, and 16384) of ground truth which can be used to precisely evaluate the completion methods at different resolutions.

To verify the effectiveness of our FBNet, we conduct a series of experiments on the MVP dataset. We train our FBNet and SnowflakeNet~\cite{xiang2021snowflakenet} on the MVP dataset and cite the results of other SOTA methods from ~\cite{pan2021variational}. The quantitative results on the high resolution (16384) completion task are shown in Table~\ref{tab:mvp}, our FBNet achieves the lowest CD distances in 14 categories. Compared with SnowflakeNet~\cite{xiang2021snowflakenet}, FBNet reduces averaged CD error with a margin of 16\%, which demonstrates the superior performance of our method.

The various resolutions completion results are shown in Table~\ref{tab:mvp_all}, our FBNet outperforms all the other methods with the lowest reconstruction error and highest F1-Score. The visualization results of 2048 points completion are shown in Fig.~\ref{fig:mvp2k}, our FBNet can not only get the lowest reconstruction errors but also recover fine-grained details of the targets. Take the second row, for example, the proposed FBNet can precisely recover the missing fuselage of the plane. PCN~\cite{yuan2018pcn}, TopNet~\cite{tchapmi2019topnet} and MSN~\cite{liu2020morphing} can only generate coarse plane shape and fail to maintain the original geometric information of partial input. VRCNet~\cite{pan2021variational} and SnowflakeNet~\cite{xiang2021snowflakenet} get better results compared to the previous method,however they also generates outliers during the completion processing.

\begin{table}[!ht]
\begin{center}
\caption{Quantitative results on PCN dataset (16384) in terms of per-point L1 Chamfer distance  $\times 10^{3}$). Our FBNet gets the lowest reconstruction error in all 8 categories.}
\scalebox{1}{
\begin{tabular}{l|cccccccc|c}
\Xhline{2\arrayrulewidth}
Methods & Airplane & Cabinet &  Car  &  Chair & Lamp & Sofa & Table & Watercraft  &  Avg.  \\
\hline
PCN \cite{yuan2018pcn} & 5.50 & 22.70 & 10.63 & 8.70 & 11.00 & 11.34 & 11.68 & 8.59 & 9.64 \\
TopNet \cite{tchapmi2019topnet}  & 7.61 & 13.31 & 10.90 & 13.82 & 14.44 & 14.78 & 11.22 & 11.12 & 12.15 \\
CRN \cite{wang2020cascaded} & 4.79 & 9.97 & 8.31 & 9.49 & 8.94 & 10.69 & 7.81 & 8.05  & 8.51 \\
GRNet \cite{xu2020grid}  & 6.45 & 10.37 & 9.45 & 9.41 & 7.96 & 10.51 & 8.44 & 8.04 & 8.83\\
PMPNet \cite{wen2021pmp}  & 5.65 & 11.24 & 9.64 & 9.51 & 6.95 & 10.83 & 8.72 & 7.25 & 8.73\\
NSFA~\cite{zhao2020exploring}  & 4.76 & 10.18 & 8.63 & 8.53 & 7.03 & 10.53 & 7.35 & 7.48 & 8.06\\
VRCNet~\cite{pan2021variational} & 4.78 & 9.96 & 8.52 & 9.14 & 7.42 & 10.82 & 7.24 & 7.49 & 8.17\\
PoinTr~\cite{yu2021pointr}  & 4.75 & 10.47 & 8.68 & 9.39 & 7.75 & 10.93 & 7.78 & 7.29 & 8.38\\
VE-PCN~\cite{wang2021voxel} & 4.80 & 9.85 & 9.26 & 8.90 & 8.68 & 9.83 & 7.30 & 7.93 & 8.32\\
SnowflakeNet \cite{xiang2021snowflakenet}  & 4.29 & 9.16 & 8.08 & 7.89 & 6.07 & 9.23 & 6.55 & 6.40 & 7.21\\
\hline
\textbf{FBNet (Ours)} & \textbf{3.99} & \textbf{9.05} & \textbf{7.90} & \textbf{7.38} & \textbf{5.82} & \textbf{8.85} & \textbf{6.35} & \textbf{6.18} & \textbf{6.94} \\
\Xhline{2\arrayrulewidth}
\end{tabular}}
\label{tab:shapenet}
\end{center}
\end{table}

\subsection{Evaluation on PCN Dataset.}

We also evaluate our FBNet with other completion methods on the PCN dataset~\cite{yuan2018pcn}. The PCN dataset is derived from ShapeNet dataset~\cite{chang2015shapenet}, which covers 30974 CAD models from 8 categories. The resolutions of partial input and ground truth are 2048 and 16384, respectively.
The quantitative results are represented in Table~\ref{tab:shapenet}. The proposed FBNet achieves the lowest CD errors in all 8 categories, which shows the robust generalization capability of our method across categories.

\begin{figure}[t]
\begin{minipage}[a]{.6\linewidth}
\begin{subfigure}
\centering
\includegraphics[scale=0.25]{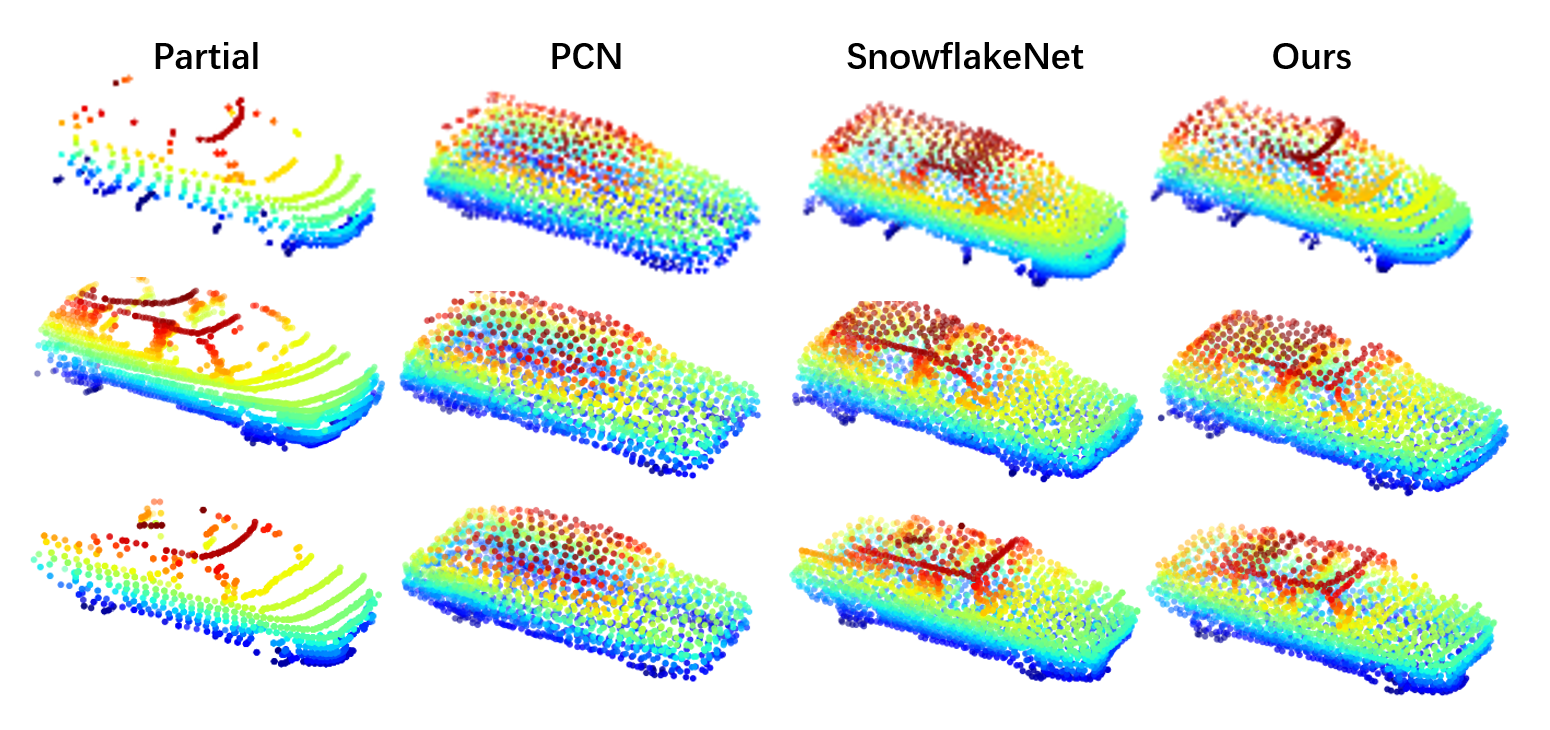}
\label{fig:kitti}
\caption{Qualitative results on KITTI Dataset}
\end{subfigure}
\end{minipage}
\begin{minipage}[b]{.4\linewidth}
\begin{center}
\captionof{table}{Quantitative results on KITTI dataset in terms of FD ($\times 10^{4}$) and MMD ($\times 10^{2}$). Lower is better.}
\scalebox{1}{

\begin{tabular}{c|cc}
\Xhline{2\arrayrulewidth}
Methods & FD$\downarrow$ & MMD$\downarrow$  \\
\hline
PCN~\cite{yuan2018pcn}  & 11.12 &  \textbf{2.50} \\
SnowflakeNet~\cite{xiang2021snowflakenet} &  2.08 & 2.81 \\
\hline
FBNet (Ours) & \textbf{0.52} & 2.97 \\
\Xhline{2\arrayrulewidth}
\end{tabular}}

\label{tab:kittifd}
\end{center}
\end{minipage}
\end{figure}




\subsection{Evaluation on KITTI Dataset.}


We further evaluate our method on KITTI~\cite{geiger2013vision} dataset, which includes 2401 real-scanned partial cars captured by a LiDAR. We directly test the models trained on the MVP dataset (2048 points) without any finetuning or retraining operations. 
As the KITTI dataset does not have ground truth shapes, we can not directly calculate reconstruction error of different methods.
The PCN~\cite{yuan2018pcn} proposed the Fidelity Distance (FD) and the Mini Match Distance (MMD) to evaluate the model's performance. We also use these two metrics in our experiment. The quantitative results in terms of FD and MMD metrics are reported in Tabel~\ref{tab:kittifd}. Our FBNet achieves lower FD compared with SnowflakeNet~\cite{xiang2021snowflakenet} and PCN~\cite{yuan2018pcn}. In addition, we visualize the reconstructed results, as shown in Fig. 7
. We can see that FBNet can not only generate the general shape of the car but also preserve the observed fine-grained details.

\section{Ablation Study}

\subsection{Feedback Refinement Mechanism}

Our ablation experiments are conducted on the MVP dataset (2048 points). We set different time steps and enable/disable the feedback connections to evaluate the effectiveness of the feedback refinement mechanism. The experiment results are shown in Table~\ref{tab:albfeedback}. When number of time steps $T$ is set to 1, our FBNet becomes a feedforward network without unfolding operation across time steps. The FBNet with 2 and 3 time steps are reported at the bottom of Table~\ref{tab:albfeedback}. Our FBNet with feedback mechanism achieves superior performance compared with the feedforward version of FBNet ($T=1$). The FBNet without feedback connections is also evaluated shown in the second and third rows of Table~\ref{tab:albfeedback}. Without the feedback connections, the performance only gets slightly improved as the feature refinement is invalid across time steps. The gain comes from the updated input of the first FBAC block across time steps.

\begin{table}[t]
\begin{center}
\caption{Ablation study of feedback mechanism on MVP dataset (2048 points).}
\scalebox{1}{
\begin{tabular}{c|c|cc}
\Xhline{2\arrayrulewidth}
Time Steps $T$ & Feedback & CD & F1 \\
\hline
1 &  & 5.41 & 0.527\\
2 &  & 5.36 & 0.526 \\
3 &  & 5.33 & 0.513 \\
\hline
2 & \checkmark & 5.19 & 0.529\\
3 & \checkmark & \textbf{5.06} & \textbf{0.532} \\
\Xhline{2\arrayrulewidth}
\end{tabular}}
\label{tab:albfeedback}
\end{center}
\end{table}

\subsection{Input initialization of FBAC}

\begin{table}[htbp]
\begin{center}
\caption{The comparison of different input initialization strategies.}
\scalebox{1.0}{
\begin{tabular}{c|cc|ccc|cc}
\Xhline{2\arrayrulewidth}
\multirow{2}{*}{Strategy} & \multicolumn{2}{c|} {First Block} & \multicolumn{3}{c|}{Others Block} & \multirow{2}{*}{CD} & \multirow{2}{*}{F1} \\ \cline{2-6}
&\quad \text{$P_{c}$+$P_{part}$} \quad\quad & \text{$P_{fb}$+$P_{part}$} & \quad \text{$P_{ff}$} \quad & \quad \text{$P_{ff}$+$P_{part}$} \quad & \quad \text{$P_{ff}$+$P_{fb}$} &  &  \\
\hline
\hline
A &\checkmark  &  & \quad\checkmark &  &  & 5.44 & 0.520 \\
B & \checkmark &  &  & \quad \checkmark &  & 5.28 &  0.528 \\
C & & \checkmark & \quad \checkmark &  &  & 5.31 &  0.508 \\
D & & \checkmark &  & & \quad \checkmark &  19.93& 0.233   \\
E(Ours) & & \checkmark &  & \quad \checkmark &  & \textbf{5.06} & \textbf{0.532}  \\
\Xhline{2\arrayrulewidth}
\end{tabular}}
\label{vis:input}
\end{center}
\end{table}

As shown in Fig.~\ref{fig:overview}, to synthesize the input of the present FBAC block, we aggregate partial and output of the previous block to a new point cloud and then downsample it via the FPS algorithm. Through this initialization operation, we refine the previous outputs with the original geometric information from partial shapes. We conduct ablation studies to evaluate the influence of different input initialization strategies. For the first FBAC block, we design two initialization ways: aggregation of HGNet and partial input $P_{c}$+$P_{part}$ across all steps, aggregation of feedback points and partial input $P_{fb}$+$P_{part}$ at $t>0$ step. For other blocks, we design three strategies: (1) Feedforward points $P_{ff}$ only. (2)Aggregation of feedforward points and partial input $P_{ff}$+$P_{part}$. (3) Aggregation of feedforward points and feedback points $P_{ff}$+$P_{fb}$. The experimental results are reported in Table~\ref{vis:input}, which proves the effectiveness of our initialization strategy.

\subsection{Adaptive Graph Pooling}

\begin{table}
\begin{center}
\caption{The comparison of different pooling methods used in HGNet on the MVP dataset (2048 points).}
\scalebox{1}{
\begin{tabular}{c|cc}
\Xhline{2\arrayrulewidth}
Pooling Methods & CD & F1 \\
\hline
Graph Pooling~\cite{zhou2021adaptive}  & 5.48 & 0.510 \\
Point Pooling~\cite{qi2017pointnet++} & 5.23 & 0.521 \\
AdaptGP (Ours) & \textbf{5.06} & \textbf{0.532} \\
\Xhline{2\arrayrulewidth}
\end{tabular}}
\label{tab:pooling}
\end{center}
\end{table}

We study the effectiveness of the proposed AdaptGP method used in HGNet. The HGNet equipped with AdaptGP gets lowest CD error and best F1-score with a large margin compared with point pooling~\cite{qi2017pointnet++}, regular graph pooling~\cite{zhou2021adaptive}, as reported in Table~\ref{tab:pooling}. The AdaptGP pooling used in HGNet can improve the average performance of our FBNet.

\section{Conclusion}

In this paper, we propose a novel feedback network for point cloud completion, named FBNet. By introducing the feedback connection in FBAC blocks, FBNet can learn more representative and informative low-level features with the help of rerouted high-level information. As the result, the FBNet gradually refines the completion results across time steps and finally gets impressive complete shapes. Exhaustive experiments on several datasets indicate that our FBNet achieves superior performance compared to state-of-the-art methods.

\clearpage
%
%
\bibliographystyle{splncs04}
\bibliography{egbib}
\clearpage

\appendix
\section{Appendix}
We provide more detailed settings and experimental results for our FBNet.
\section{Detailed Settings}
\subsection{Network Implementation Details}

{\bf HGNet.}
The hierarchical graph-based encoder stacks 3 EdgeConv and 2 AdaptGP layers. Both EdgeConv and AdaptGP use $k$-nearest neighbors (kNN) as the grouping operation, where we set $k=16$ throughout the paper. The $k$ of kNN strategy in our cross transformer is set to 16.

We use 3 full-connected layers ($1024\times1024, 1024\times1024,1024\times128*3$) to decode the global feature to coarse shape $P_{c}$ with $N_{c}=128$. Besides, we aggregate $P_{c}$ and the partial input, and downsample it to a new coarse output $P_{c}^{'}$ with size $512\times3$, which is the input of $0$-th FBAC block at first time step($t=0$).

{\bf FBAC block.}
The FBAC block in the FBNet is a lightweight sub-network and its weight parameters are shared across time steps.
For feature extraction, feedback exploitation and feature expansion modules, channel dimensions of these modules' output features are set to 128. The NodeShuffle layer~\cite{Qian2021PUGCNPC} is used to expand features to higher resolution ones. The EdgeConv and MLPs are used to expand input features by $r$ times on channel dimension, and then a shuffle operation is used to rearrange the feature map.

{\bf Time Steps and Up-sampling Ratios.}
We set the input size of first FBAC block to $512\times3$, and the upsampling ratio in each FBAC block is set to expand point clouds and generate fine-grained shapes.
The detailed upsampling ratios of different resolution completion task is shown in Table~\ref{tab:upratio}.
We set $T=3$ for 2048 points completion task, and for other higher resolution tasks, we set $T=2$ to shorten training time and get competitive inference cost compared with recent SOTA works.


\subsection{Training details}
We implement our method with PyTorch and use the Adam optimizer \cite{kingma2014adam} with $\beta_1$ = 0.9 and $\beta_2$ = 0.999 to train it.

{\bf MVP Dataset.} For MVP dataset, we set initial learning rate to $10^{-3}$ with a decay of 0.1 every 30 epochs for our method.  We train our method on NVIDIA V100 16G GPU with batch size 48 and take 100 epochs to converge.

{\bf PCN Dataset.} In PCN dataset, there are 8 incomplete shapes that
are captured from 8 different views for each object. In each training epoch, we use all 8 views data for our method as PCN did~\cite{yuan2018pcn}. We set initial learning rate to $10^{-3}$ with a decay of 0.7 every 16 epochs for our method. We train our method on NVIDIA P40 24G GPU with batch size 80 and take 100 epochs to converge.

\begin{table}[t]
\begin{center}
\caption{The number of time steps and upsampling ratios for various resolution completion tasks in our FBNet.}
\scalebox{1}{
\begin{tabular}{c|c|cc}
\Xhline{2\arrayrulewidth}
Resolutions & Time Steps ($T$) & Up-sampling Ratios of FBAC blocks \\
\hline
2048 & 3 & 1,2,2\\
4096 & 2 & 1,2,4 \\
8192 & 2 & 1,2,8 \\
16384 & 2 & 1,2,16 \\
\Xhline{2\arrayrulewidth}
\end{tabular}}
\label{tab:upratio}
\end{center}
\end{table}

\begin{table}[t]
\begin{center}
\caption{The effect of feedback refinement mechanism on MVP dataset (2048 points). The FBNet ($T=3$) is unfolded across time and the output of each time step is evaluated.}
\scalebox{1}{
\begin{tabular}{c|c|cc}
\Xhline{2\arrayrulewidth}
$t$-th Time Step & CD & F1 \\
\hline
0 & 5.79 & 0.510\\
1 & 5.17 & 0.527 \\
2 & 5.06 & 0.530 \\
\Xhline{2\arrayrulewidth}
\end{tabular}}
\label{tab:unfolding}
\end{center}
\vspace{-2em}
\end{table}

\begin{table}[t]
\begin{center}
\caption{ The time and space complexity on MVP (16K) dataset.}
\scalebox{1}{
\begin{tabular}{c|cccccc}
\Xhline{2\arrayrulewidth}
models & PCN & VRCNet & SnowflakeNet & PoinTr &  FBNet($T=1$) & FBNet($T=2$) \\
\hline
Params (M) & 6.86 & 16.30 & 19.32 &  31.27 & 4.96 & 4.96\\
Time (ms) & 0.63 & 29.17 & 3.47 & 4.98 & 2.81 & 4.58\\
CD        & 6.02 & 3.06 & 2.73 & 3.74 & 2.59 & 2.29\\
\Xhline{2\arrayrulewidth}
\end{tabular}}
\label{tab:modelsize}
\end{center}
\vspace{-2em}
\end{table}

\section{More Results}
\subsection{The effectiveness of feedback refinement mechanism}
\begin{figure}[t]
\centering
\scalebox{1.0}{
\includegraphics[width=\textwidth]{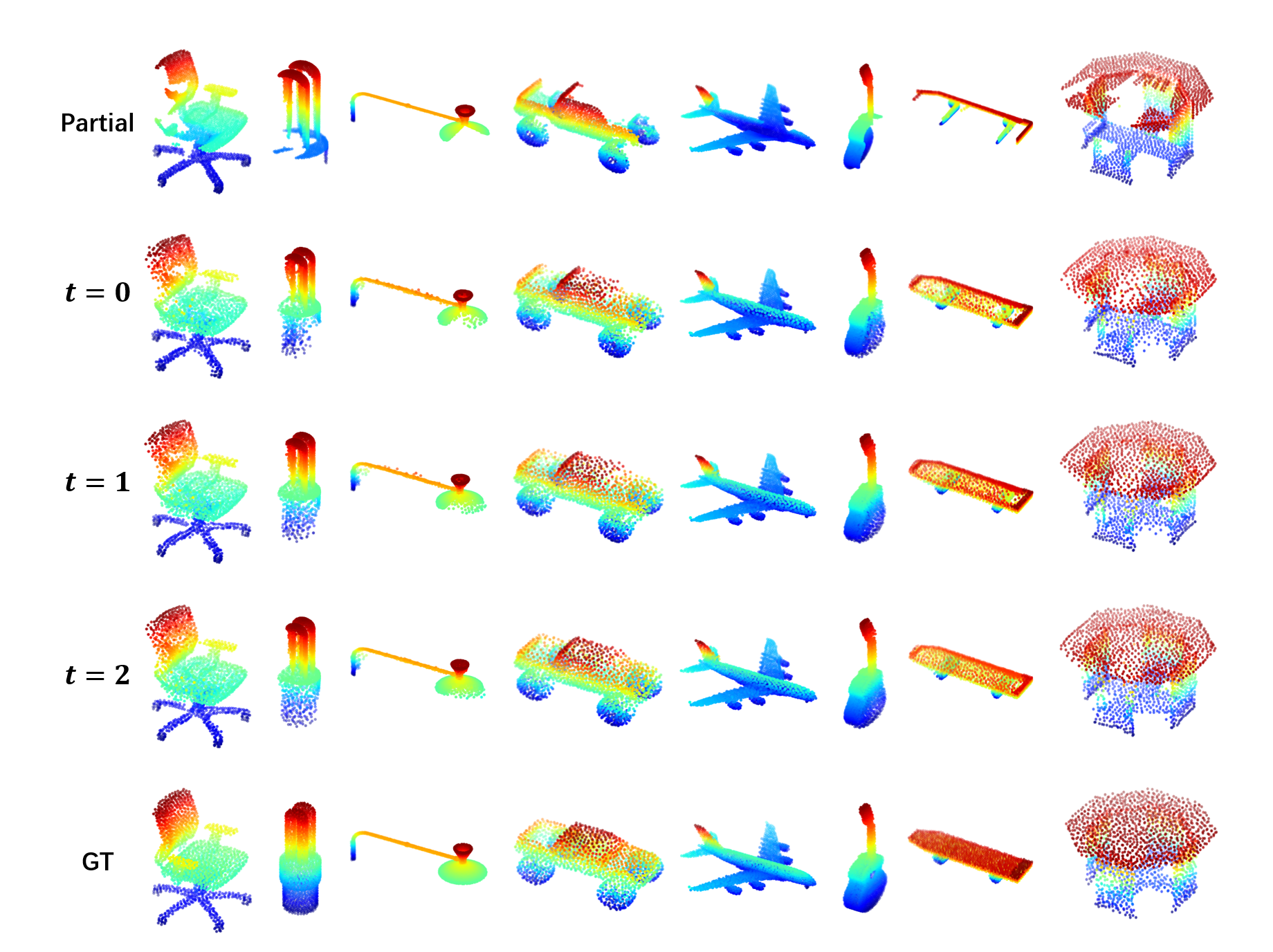}}
\caption{Qualitative comparison of different time step results of our FBNet. The output is refined gradually across time steps via feedback refinement mechanism.}
\label{fig:unfolding}
\vspace{-2em}
\end{figure}

We unfold our trained model across time steps and the output of each time step is evaluated. Quantitative results is reported in Table~\ref{tab:unfolding}, the outputs at present step $t$ gets lower reconstruction error compared to the ones at previous time step $t-1$. The qualitative comparison of different time step is shown in Fig~\ref{fig:unfolding}, FBNet gradually refines the completion results across time steps via feedback refinement mechanism. Take the third column for example, the completed lamp becomes more complete and less noise with the increasing of time step $t$. Therefore, our FBNet has the ability to train once and dynamically adjust time step $t$ to balance model inference efficiency and effectiveness for various devices with different computational resources.

\subsection{ The time and space complexity}

We compare our FBNet with other methods in terms of parameter size and inference cost on MVP dataset (16384 points). We test the infer time on NVIDIA TITAN X 12G GPU with batchsize 32, the results are shown in Table~\ref{tab:modelsize}. The FBAC block in the FBNet is a lightweight sub-network and its weight parameters are shared across time steps. Although 3 FBACs are stacked and will be expanded in the time dimension, our FBNet still has the smallest parameter size and competitive inference
cost compared with recent SOTA works.


\subsection{More visualization results}

\begin{figure}[t]
\centering
\scalebox{1.0}{
\includegraphics[width=\textwidth]{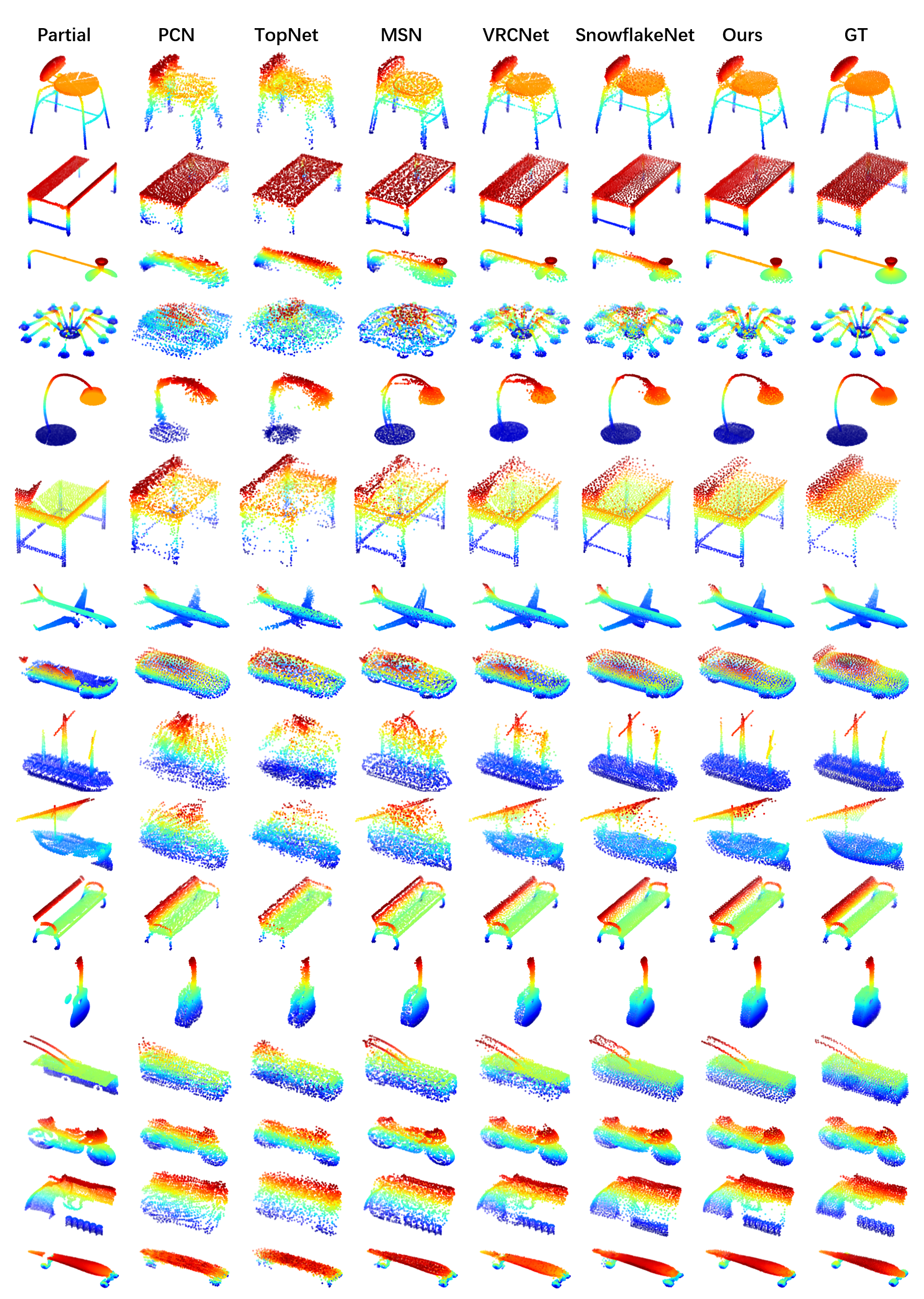}}
\caption{More qualitative comparison of different methods on MVP dataset(2048 points).}
\label{fig:mvp2k_sup}
\end{figure}

In Fig~\ref{fig:mvp2k_sup}, we provide more shape completion results
on MVP dataset. FBNet has the ability to recover fine-grained details of targets with less noise, especially under challenging categories such as lamp and watercraft.

\end{document}